\pdfoutput=1
\documentclass[conference]{IEEEtran}
\usepackage{cite}
\usepackage{booktabs}
\usepackage{multirow}
\usepackage{siunitx}
\usepackage{subcaption}
\usepackage{multicol}
\usepackage{url}
\usepackage[pdftex]{graphicx}
\usepackage[export]{adjustbox}
\usepackage{algorithm,algorithmic}
\usepackage{lettrine}
\usepackage [fleqn] {amsmath}
\usepackage{graphicx}
\usepackage{mathptmx}
\usepackage[T1]{fontenc}
\usepackage{colortbl}
\usepackage{color}
\usepackage{rotating}
\usepackage{float}
\usepackage{caption}
\usepackage{hhline}
\definecolor{maroon}{cmyk}{0,0.87,0.68,0.32}
\usepackage{cases}
\usepackage{amsfonts}

\usepackage{blindtext}
\usepackage{bm}
\usepackage{amsmath}
\usepackage[switch]{lineno}
\usepackage{xcolor}
\usepackage{pifont}
\newcommand\amend[1]{\textcolor{blue}}

\newlength\myindent
\newcommand\bindent{%
  \begingroup
  \setlength{\itemindent}{\myindent}
  \addtolength{\algorithmicindent}{\myindent}
}
\newcommand\eindent{\endgroup}
\newcommand{\NA}{---}
\usepackage{tabularx}
\usepackage{eso-pic}
\usepackage{textcomp}
\begin{document}

\AddToShipoutPicture*{\small \sffamily\raisebox{1.2cm}{\hspace{1.8cm}978-1-7281-0397-6/20/\$31.00 \textcopyright~2020 IEEE}}
\title{Training Progressively Binarizing Deep Networks Using FPGAs}

\author{\IEEEauthorblockN{Corey Lammie, Wei Xiang, and Mostafa Rahimi Azghadi}
	\IEEEauthorblockA{College of Science and Engineering, James Cook University, Queensland 4814, Australia\\
		Email:\{corey.lammie, mostafa.rahimiazghadi, wei.xiang\}@jcu.edu.au}}
\maketitle

\begin{abstract}

While hardware implementations of inference routines for Binarized Neural Networks (BNNs) are plentiful, current realizations of efficient BNN hardware training accelerators, suitable for Internet of Things (IoT) edge devices, leave much to be desired. Conventional BNN hardware training accelerators perform forward and backward propagations with parameters adopting binary representations, and optimization using parameters adopting floating or fixed-point real-valued representations--requiring two distinct sets of network parameters.
In this paper, we propose a hardware-friendly training method that, contrary to conventional methods, progressively binarizes a singular set of fixed-point network parameters, yielding notable reductions in power and resource utilizations. We use the Intel FPGA SDK for OpenCL development environment to train our progressively binarizing DNNs on an OpenVINO FPGA. We benchmark our training approach on both GPUs and FPGAs using CIFAR-10 and compare it to conventional BNNs.
\end{abstract}


\section{Introduction}

\begin{figure*}[!t]
	\centering
	\includegraphics[width=1\textwidth]{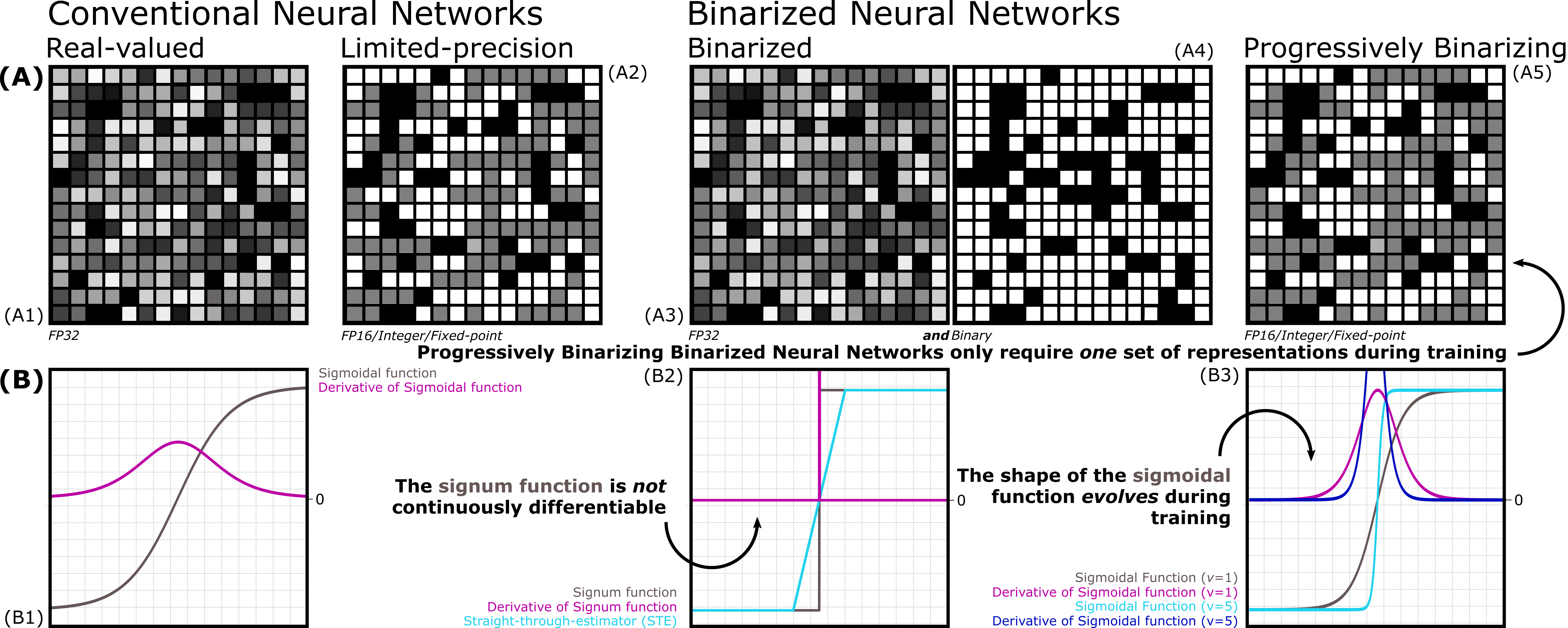}
	\caption{Depiction of (A) network parameter representations and (B) binarization and activation functions required during training for various DNNs and BNNs. Different levels of discretization are depicted using various shade palettes. DNNs require one set of real-valued (A1) or limited-precision parameters (A2) and typically have continuously differentiable activation functions (B1). In addition to the real-valued parameters (A3), BNNs also require another set of binarized parameters (A4), for which they use a STE (B2) to determine gradients of the signum function, which is not continuously differentiable. In contrast to BNNs, PBNNs require one set of real-valued parameters (A5) and use a continuously differentiable activation and binarization function, with a shape that progressively evolves during training (B3).}
	\label{im2col}
\end{figure*}

\lettrine{B}{inarization} has been used to augment the performance of Deep Neural Networks (DNNs), by quantizing network parameters to binary states, replacing many resource-hungry multiply-accumulate operations with simple accumulations~\cite{DBLP:journals/corr/CourbariauxB16}. It has been demonstrated that Binarized Neural Networks (BNNs) implemented on customized hardware can perform inference faster than conventional DNNs on state-of-the-art Graphics Processing Units (GPUs)~\cite{7929192,8693488}, while offering notable improvements in power consumption and resource utilizations~\cite{Yang:2018:FOB:3218603.3218615,Liang2018,8884910}. However, there is still a performance gap between DNNs and conventional BNNs~\cite{DBLP:journals/corr/abs-1812-11800}, which binarize parameters deterministically or stochastically. Moreover, the training routines of conventional BNNs are inherently unstable~\cite{AAAI1714619}.

During backward propagations of conventional BNN training routines, gradients are approximated using a Straight-Through Estimator (STE) as the signum function is not continuously differentiable~\cite{DBLP:journals/corr/CourbariauxB16}. The gap in performance, and the general instability of conventional BNNs compared to DNNs, can be largely attributed to the lack of an accurate derivative for weights and activations in BNNs, which creates a mismatch between binary and floating- or fixed-point real-valued representations~\cite{Lin2017}.

The training routines of DNNs that utilize continuously differentiable and adjustable functions in place of the signum function, which we denote Progressively Binarizing NNs (PBNNs), transform a complex and non-smooth optimization problem into a sequence of smooth sub-optimization problems. Such training routines that progressively binarize network parameters, were first used to binarize the last layer of DNNs to yield significant multimedia retrieval performance on standard benchmarks~\cite{DBLP:journals/corr/CaoL0Y17}. Since, various works have detailed training routines of complete PBNNs~\cite{8461456,DBLP:journals/corr/abs-1902-00730,DBLP:journals/corr/abs-1806-02988}.
However, efficient customized hardware implementations of PBNNs are yet to be explored.

In this paper, we use the Intel FPGA SDK for OpenCL development environment to implement and train novel and scalable PBNNs on an OpenVINO FPGA, which progressively binarize a singular set of fixed-point network parameters. We compare our approach to conventional BNNs and benchmark our implementations using CIFAR-10~\cite{krizhevsky2009learning}.
Our specific contributions are as follows:

\begin{itemize}
    \item We implement and present the first PBNNs using customized hardware and fixed-point number representations;
    \item We use a Piece-Wise Linear (PWL) function for binarization and activations with a constant derivative to simplify computations;
    \item We demonstrate compared to training BNNs deterministically or stochastically on CIFAR-10, PBNNs yield a marginal, yet consistent, increase in classification accuracy, and decrease both resource and power utilizations.
\end{itemize}

\section{Preliminaries}
\subsection{Conventional BNNs}
The training routines of conventional BNNs binarize parameters either deterministically or stochastically after performing parameter optimizations. Deterministic binarization is performed as per Eq. (\ref{det_binarization}).

\begin{equation}\label{det_binarization}
\mathbf{\theta_b} = \left\{\begin{array}{lr}
-1 & \textnormal{if } \mathbf{\theta} \leq 0\\
+1 & \textnormal{if } \mathbf{\theta} > 0,
\end{array}\right.
\end{equation}
where $\mathbf{\theta_b}$ denotes binarized parameters and $\mathbf{\theta}$ denotes real-valued full-precision parameters. Stochastic binarization is performed as per Eq. (\ref{stochastic_bin}), where $\sigma$ is the hard sigmoid function described in Eq. (\ref{hard_sigmoid}).

\begin{equation}\label{stochastic_bin}
\mathbf{\theta_b} = \left\{\begin{array}{ll}
+1 & \textnormal{with probability } \rho = \sigma(\mathbf{\theta}),\\
-1 & \textnormal{with probability } 1 - \rho
\end{array}\right.
\end{equation}

\begin{equation}\label{hard_sigmoid}
\sigma(\mathbf{\theta}) = \textnormal{clip} (\frac{\mathbf{\theta}+1}{2},0,1) = \textnormal{max} (0, \textnormal{min} (1,\frac{\mathbf{\theta}+1}{2}))
\end{equation}

During backward propagations, large parameters are clipped using $t_{\textnormal{clip}}$, as per Eq. (\ref{STE}), where $J$ denotes the objective function.

\begin{equation}\label{STE}
  \frac{\partial J}{\partial \mathbf{\theta}} = \frac{\partial J}{\partial \mathbf{\theta_b}} 1_{|\mathbf{\theta}| \leq t_{\textnormal{clip}}}
\end{equation}

\subsection{Progressively Binarizing DNNs}
PBNNs use a set of constrained real-valued parameters $\mathbf{\theta_l}$ at each layer $\mathbf{l}$, which are not directly learnable, but are a function of learnable parameters $\mathbf{P}$ at each layer~\cite{8461456,DBLP:journals/corr/abs-1902-00730}. The shape of $\theta(P)$ evolves during training, to closely resemble the signum function 
once training is complete. The hyperbolic tangent function is commonly used to relate $\theta$ and $P$, as described in Eq. (\ref{weights_parameters}).

\begin{equation}\label{weights_parameters}
	\theta (P) = \textnormal{tanh}(v \cdot P),
\end{equation}
where $v$ is an adjustable scale parameter, which is used to evolve the shape of Eq. (\ref{weights_parameters}). The derivative of Eq. (\ref{weights_parameters}) is described in Eq. (\ref{weights_parameters_dydx}).

\begin{equation}\label{weights_parameters_dydx}
	\frac{\partial \textnormal{tanh}(v \cdot P)}{\partial P} = v \cdot (1 - \textnormal{tanh}^2(v \cdot P)).
\end{equation}

\begin{figure}[!b]
	\centering
	\includegraphics[width=0.4\textwidth]{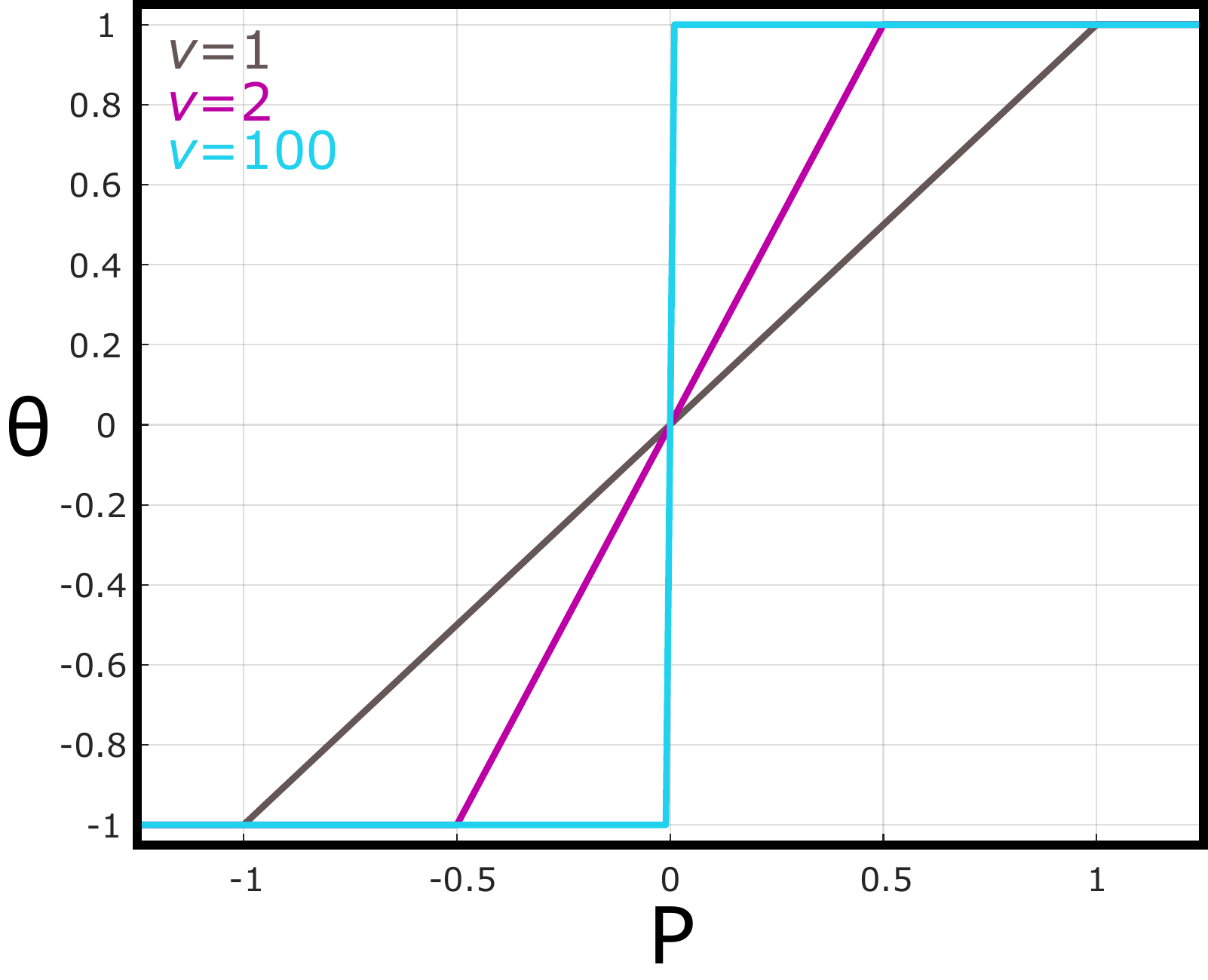}
	\caption{Activation and binarization functions used for our progressive binarization training routine.}
	\label{activation_binarization}
\end{figure}

As $v$ increases, the shape of Eq. (\ref{weights_parameters}) better mimics that of the signum function. During training, parameters, denoted using $\mathbf{P}$, are optimized to minimize a loss function, while $v$ is progressively increased. After training is completed, $v$ is sufficiently large that the parameters, $\mathbf{\theta}$, are very close to $\in [-1 ,1]$, as depicted in Fig. \ref{im2col}. The final binary parameters can simply be obtained by passing $\mathbf{\theta}(\mathbf{P})$ through the signum function.

\begin{algorithm}[!b]
	\caption{The training rotuine adopted by all of our progressively-binarizing DNNs.}
	\begin{algorithmic}
		\renewcommand{\algorithmicrequire}{\textbf{Input:}}
		\renewcommand{\algorithmicensure}{\textbf{Output:}}
		\REQUIRE Network hyperparameters (the learning rate schedule, $\mathbf{\eta}$, scale parameter schedule, $\mathbf{v}$, batch size, $\Im$, gradient optimizer, loss function, $\mathbf{J}(\mathbf{\theta}, \mathbf{y_i'}(a_{-1}), \mathbf{y_i})$, and the number of training epochs).
		\ENSURE Trained binary weights and biases, $\mathbf{\theta_b}$.
		\FOR{each training epoch}
  		\STATE \textbf{1. Forward Propagation}
  			\bindent
  				\STATE $\eta, v = \mathbf{\eta}[\textnormal{epoch}], \mathbf{v}[\textnormal{epoch}]$
  				\FOR{each training batch}
  					\FOR{each layer}
  						\STATE Determine $\mathbf{a_l}[n] = \mathbf{\theta_l}[n-1](\mathbf{P_l}[n-1])$
  					\ENDFOR
  				\ENDFOR
  			\eindent

        \STATE \textbf{2. Backward Propagation}
        \bindent
          \STATE Determine $\mathbf{J}(\mathbf{\theta}, \mathbf{y_i'}(a_{-1}), \mathbf{y_i})$
          \FOR{all other layers}
            \STATE Determine $\frac{\partial \mathbf{J}}{\partial \mathbf{a_{l-1}}[n]}$ using $\frac{\partial \mathbf{J}}{\partial \mathbf{a_{l}}[n]}$ and $\mathbf{\theta_l}[n-1]$
          \ENDFOR
        \eindent
        \STATE \textbf{3. Parameter Optimization}
        \bindent
          \FOR{each layer}
            \STATE Determine $\frac{\partial \mathbf{J}}{\partial \mathbf{\theta_l}[n-1]}$ using $\frac{\partial \mathbf{J}}{\partial \mathbf{a_{l}}[n]}$
            \STATE Determine $\mathbf{\theta}[n]$ using $\frac{\partial \mathbf{J}}{\partial \mathbf{\theta_l}[n-1]}$ and $\eta$
          \ENDFOR
        \eindent
		\ENDFOR
    \STATE \textbf{4. Determine the Trained Binary Parameters}
    \bindent
      \STATE $\mathbf{\theta_b} = [\,]$
      \FOR{each layer}
        \STATE $\mathbf{\theta_b} = \textnormal{concat(}\mathbf{\theta_b}, \textnormal{sign}(\mathbf{\theta_l}))$
      \ENDFOR
    \eindent
  \end{algorithmic} \label{prog_train_alg}
\end{algorithm}

\section{Implementation Details}
\subsection{Our Progressive Binarization Training Routine}
We employ a PWL function to approximate the hyperbolic tangent function, described in Eq. (\ref{pwl}) and depicted in Fig. \ref{activation_binarization} to simplify computations. In addition to reducing a non-linear function to a linear function, the derivative of Eq. (\ref{pwl}), when bounded, is constant and does not depend on $P$. Consequently, when all activations are computed simultaneously, the output of each layer during forward propagations, $\mathbf{a_l}[n]$, does not need to be stored in memory to determine gradients during backward propagations.

\begin{equation}\label{pwl}
	\mathbf{\theta (P)} = \left\{\begin{array}{lr}
	-1 & \textnormal{if } P < -1\\
	v \cdot P & \textnormal{if } -1 \leq P \leq 1\\
	+1 & \textnormal{if } P  > 1
	\end{array}\right.
\end{equation}

Algorithm \ref{prog_train_alg} provides a high-level overview of our progressive binarization training routine. Trained binary parameters can be computed after each training epoch to determine performance on the test set during training. Here, $\mathbf{a_l}[n]$ denotes the output of the $l$th layer at the $n$th epoch. As $v \gg 1$ the sign of the output of Batch Normalization (BN) is reformulated to reduce computation as per Eq. (\ref{bn})~\cite{DBLP:journals/corr/abs-1902-00730}.

\begin{equation}\label{bn}
  \textnormal{sign}(\mathbf{a_l}[n]) = \textnormal{XNOR}(I > T, \gamma > 0),
\end{equation}
where $T$ is defined in Eq. (\ref{t}). $I$ denotes the input, $\beta$ and $\gamma$ are parameters that define an affline transform, and $u_r$ and $\sigma_r$ are the running mean and standard deviation of the feature maps that pass through them.

\begin{equation}\label{t}
  T = \mu_r - \frac{\sigma_r \cdot \beta}{\gamma}
\end{equation}

We trained all networks until improvement on the test set was negligible (for 50 epochs) with a batch size $\Im = 8$. This is the largest possible batch size that makes comparison across devices possible. The initial learning rate was $\eta = 1e-3$, which was decayed by an order of magnitude every 20 training epochs, i.e. when $\textnormal{mod}(\eta, 20) = 0$.

During training, each network's scale parameter, $v$, was increased logarithmically, from 1, at the first epoch, to 1000, at the final epoch. Eq. (\ref{bn}) was used to determine the output of all batch normalization layers when $v \geq 500$. Adam~\cite{Kingma2014AdamAM} was used to optimize network parameters and Cross Entropy (CE)~\cite{DBLP:journals/corr/abs-1805-07836} was used to determine network losses. After the trained binary parameters were determined, for all our implementations, a conventional OpenCL BNN inference accelerator was used to perform inference on the CIFAR-10 test set.

\subsection{Network Architecture}
The network architecture, previously used in~\cite{pmlr-v38-lee15a}, was used in all of our DNNs. This architecture is a variant of the VGG~\cite{Simonyan2014VeryDC} family of network architectures. It is summarized in Table \ref{network_architecture}. For each convolutional and pooling layer, $f$ denotes the number of filters, $k$ determines the filter size, $s$ is the stride length, and $p$ denotes the padding. Here, $N$ is the number of output neurons for each fully connected layer. All convolutional and fully connected layers are sequenced with batch normalization and activation layers. The last fully connected layer adopts real-valued representations.

\begin{table}[!b]
\centering
\caption{Adopted Network Architecture.}
\begin{adjustbox}{width=0.5\textwidth}
  \begin{tabular}{lrc}
\toprule
 \textbf{Layer}  & \textbf{Output Shape}  & \textbf{Binarized} \\
\midrule
 Convolutional,~$f = 128, k=3, s=1, p=1$  & $(128 \times 32 \times 32)$  & \ding{51} \\
Convolutional,~$f = 128, k=3, s=1, p=1$  & $(128 \times 32 \times 32)$  & \ding{51} \\
Max Pooling,~$k=2, p=2$  & $(128 \times 16 \times 16)$  & \ding{51} \\
Convolutional,~$f = 128, k=3, s=1, p=1$  & $(128 \times 16 \times 16)$  & \ding{51} \\
Convolutional,~$f = 256, k=3, s=1, p=1$  & $(256 \times 16 \times 16)$  & \ding{51} \\
Max Pooling,~$k=2, p=2$  & $(256 \times 8 \times 8)$  & \ding{51} \\
Convolutional,~$f = 256, k=3, s=1, p=1$  & $(256 \times 8 \times 8)$  & \ding{51} \\
Convolutional,~$f = 512, k=3, s=1, p=1$  & $(512 \times 8 \times 8)$  & \ding{51} \\
Max Pooling,~$k=2, p=2$  & $(512 \times 4 \times 4)$  & \ding{51} \\
Fully Connected, $N = 1024$  & $(1024)$  & \ding{51} \\
Fully Connected,~$N = 1024$  & $(1024)$  & \ding{51} \\
Fully Connected,~$N = 10$  & \begin{tabular}[c]{@{}r@{}}$(10)$\\ \end{tabular} &  \\
\bottomrule
\end{tabular}
\end{adjustbox}\label{network_architecture}
\end{table}

\begin{table*}[!t]
\centering
\caption{Implementation results obtained using the CIFAR-10 dataset for GPU and FPGA accelerated networks. $^1$The mean and standard deviations reported for the Training Time per Epoch (s) metric are determined over 50 training epochs. $^2$Similarly to conventional networks, the unbounded ReLU~\cite{pmlr-v15-glorot11a} activation function was used instead of Eq. (\ref{pwl}) for the real-valued FP-32 baseline implementation on GPU. $^3$The same test set accuracy was achieved for GPU and FPGA implementations.}
\begin{adjustbox}{width=1\textwidth}
\begin{tabular}{lrrrrrrrr}
\toprule
\multicolumn{1}{c}{\multirow{2}{*}{ \textbf{Training Routine} }} & \multicolumn{2}{c}{\textbf{Total Kernel Power Usages (W)} } & \multicolumn{2}{c}{\textbf{Total Training Time (s)} } & \multicolumn{2}{c}{\textbf{Training Time per Epoch (s)}$^1$ } & \multicolumn{2}{c}{\textbf{Test Set Accuracy (\%)$^3$} } \\
\multicolumn{1}{c}{} & \textbf{FPGA}  & \textbf{GPU}  & \textbf{FPGA}  & \textbf{GPU}  & \textbf{FPGA}  & \textbf{GPU}  & \textbf{FPGA}  & \textbf{GPU}  \\
\midrule
\multicolumn{9}{c}{8-bit Fixed Point} \\
\midrule
Stochastic & 8.06 & 133.9 & 1,592.73 & 2,613,67 & $31.85 \pm 0.21$ & $52.27 \pm 0.37$ & 85.91 & 85.91\\
Deterministic & 7.95 & 133.0 & 1,523.17 & 2,497.72 & $30.46 \pm 0.18$ & $49.95 \pm 0.31$ & 85.56 & 85.56\\
Progressive & 7.60 & 130.3 & 1,383.17 & 2,315.97 & $27.66 \pm 0.17$ & $46.31 \pm 0.32$ & 86.28 & 86.28 \\
\midrule
\multicolumn{9}{c}{16-bit Fixed Point} \\
\midrule
Stochastic & 10.19 & 134.2 & 1,989.25 & 3,147.23 & $39.78 \pm 0.17$ & $62.94 \pm 0.31$ & 86.45 & 86.45\\
Deterministic & 10.03 & 132.8 & 1,907.17 & 2,909.62 & $38.14 \pm 0.19$ & $58.19 \pm 0.36$ & 86.16 & 86.16\\
Progressive & 9.27 & 130.5 & 1,729.32 & 2,685.22 & $34.58 \pm 0.22$ & $53.70 \pm 0.34$ & 86.94 & 86.94\\
\midrule
\multicolumn{9}{c}{FP32 Baseline} \\
\midrule
Real-valued$^2$  & \NA & 137.1 & \NA & 2,524.20 & \NA & $50.48 \pm 0.35$ & \NA & 86.77 \\
\bottomrule
\end{tabular}
\end{adjustbox}\label{bigtable}
\end{table*}

\begin{table}[!b]
	\centering
	\caption{Comparison of device FPGA utilization for various binarization training approaches. The numbers are extracted from $acl\_quartus\_report.txt$, generated by Quartus Prime Design Suite 18.1.}
\begin{tabularx}{0.5\textwidth}{lrrr}
\toprule
\textbf{Training Routine}  & \multicolumn{1}{c}{\textbf{Deterministic }} & \multicolumn{1}{c}{\textbf{Stochastic }} & \multicolumn{1}{c}{\textbf{Progressive }} \\
\midrule
Device & \multicolumn{3}{c}{Intel FPGA OpenVINO} \\
Dataset & \multicolumn{3}{c}{CIFAR-10} \\
\midrule
\multicolumn{4}{c}{8-bit Fixed Point} \\
\midrule
Flip Flops (\%) & 63.19 & 66.42 & 62.95 \\
ALMs (\%) & 81.38 & 84.87 & 76.92 \\
DSPs (\%) & 100.00 & 100.00 & 93.20 \\
\midrule
\multicolumn{4}{c}{16-bit Fixed Point} \\
\midrule
Flip Flops (\%) & 96.06 & 98.43 & 91.96 \\
ALMs (\%) & 90.40 & 94.31 & 85.54 \\
DSPs (\%) & 100.00 & 100.00 & 100.00 \\
\bottomrule
\end{tabularx}
\label{comparisons_t}
\end{table}

\subsection{Hardware Architecture}
All of our implementations are described using the heterogeneous OpenCL~\cite{5457293} framework, in which multiple OpenCL kernels are accelerated using either FPGAs or GPUs that are controlled using C++ host controllers. For FPGA implementations a SoC is used as the host controller, whereas for GPU implementations a CPU is used. We note that the power consumption of our FPGA implementations could be further decreased by realizing them using Hardware Description Language (HDL), removing the host controller, however, this would make fair comparisons between GPU and FPGA implementations difficult~\cite{Sorensen:2016:HGC:2909437.2909440}.

\section{Implementation Results}
In order to investigate the performance of our progressively binarizing training routine, CIFAR-10 was used. Prior to training, the color channels of each image were normalized using mean and standard deviation values of (0.4914, 0.2023), (0.4822, 0.1994), and (0.4465, 0.2010), for the red, green, and blue image channels, respectively. This normalization was performed because it has demonstrated significant performance on the ImageNet dataset~\cite{Krizhevsky:2017:ICD:3098997.3065386}. We compare FPGA implementations adopting 16-bit and 8-bit fixed-point real-valued representations, as a large degradation in performance was observed when using smaller bit widths.

To compile OpenCL kernels for the OpenVINO FPGA, the Intel FPGA SDK for OpenCL Offline Compiler (IOC) was used, as part of the Intel FPGA SDK for OpenCL and Quartus Prime Design Suite 18.1. For our GPU implementations, a Titan V GPU was used to execute OpenCL kernels and an AMD Ryzen 2700X @ 4.10 GHz Overclocked (OC) CPU was used to drive the host controller. We used version 430.50 of the Titan V GPU driver to launch compute kernels. We report all GPU and FPGA implementation results in Table \ref{bigtable}.

From Table \ref{bigtable}, it can be observed that our progressive training routine consumed the least power and had the smallest total training time on FPGA. Moreover, when adopting 16-bit fixed-point real-valued representations during training it achieved the largest test set accuray. We believe that, similarly to~\cite{DBLP:journals/corr/CourbariauxB16}, this can be attributed to the additional regularization that binarized parameters introduce. We note that the total training times of our GPU and FPGA implementations are not indicative of those with larger batch sizes, and that the available resources on the FPGA used, restricted us to use $\Im = 8$ across all devices.

The device utilization of our FPGA implementations is presented in Table \ref{comparisons_t}. Our progressive binarizing training routine consumes notably less Adaptive Logic Modules (ALMs) and Flip Flops than deterministic and stochastic routines for both 16- and 8-bit fixed-point representations. Digital Signal Processor (DSP) utilization is similar to deterministic and stochastic routines, and is only decreased marginally when 8-bit fixed-point real-valued representations are adopted.

\section{Conclusion}
We proposed and implemented novel and scalable PBNNs on GPUs and FPGAs. We compared our approach to conventional BNNs and real-valued DNNs using GPUs and FPGAs and demonstrated notable reductions in power and resource utilizations for CIFAR-10. This was achieved through approximations and hardware optimizations, as well as using only one set of network parameters compared to conventional BNNs. We leave further hardware-level dissemination, upscaling, hyperparameter optimization, and tuning to future works.
\newpage

\bibliographystyle{IEEEtran}
\bibliography{References}

\begin{thebibliography}{10}
\providecommand{\url}[1]{#1}
\csname url@samestyle\endcsname
\providecommand{\newblock}{\relax}
\providecommand{\bibinfo}[2]{#2}
\providecommand{\BIBentrySTDinterwordspacing}{\spaceskip=0pt\relax}
\providecommand{\BIBentryALTinterwordstretchfactor}{4}
\providecommand{\BIBentryALTinterwordspacing}{\spaceskip=\fontdimen2\font plus
\BIBentryALTinterwordstretchfactor\fontdimen3\font minus
  \fontdimen4\font\relax}
\providecommand{\BIBforeignlanguage}[2]{{%
\expandafter\ifx\csname l@#1\endcsname\relax
\typeout{** WARNING: IEEEtran.bst: No hyphenation pattern has been}%
\typeout{** loaded for the language `#1'. Using the pattern for}%
\typeout{** the default language instead.}%
\else
\language=\csname l@#1\endcsname
\fi
#2}}
\providecommand{\BIBdecl}{\relax}
\BIBdecl

\bibitem{DBLP:journals/corr/CourbariauxB16}
\BIBentryALTinterwordspacing
M.~Courbariaux and Y.~Bengio, ``{BinaryNet: Training Deep Neural Networks with
  Weights and Activations Constrained to +1 or -1},'' \emph{CoRR}, vol.
  abs/1602.02830, 2016. [Online]. Available:
  \url{http://arxiv.org/abs/1602.02830}
\BIBentrySTDinterwordspacing

\bibitem{7929192}
E.~{Nurvitadhi}, D.~{Sheffield}, {Jaewoong Sim}, A.~{Mishra}, G.~{Venkatesh},
  and D.~{Marr}, ``{Accelerating Binarized Neural Networks: Comparison of FPGA,
  CPU, GPU, and ASIC},'' in \emph{2016 International Conference on
  Field-Programmable Technology (FPT)}, Dec 2016, pp. 77--84.

\bibitem{8693488}
C.~{Lammie}, A.~{Olsen}, T.~{Carrick}, and M.~{Rahimi Azghadi}, ``{Low-Power
  and High-Speed Deep FPGA Inference Engines for Weed Classification at the
  Edge},'' \emph{IEEE Access}, vol.~7, pp. 51\,171--51\,184, 2019.

\bibitem{Yang:2018:FOB:3218603.3218615}
\BIBentryALTinterwordspacing
L.~Yang, Z.~He, and D.~Fan, ``{A Fully Onchip Binarized Convolutional Neural
  Network FPGA Impelmentation with Accurate Inference},'' in \emph{Proceedings
  of the International Symposium on Low Power Electronics and Design}, ser.
  ISLPED '18.\hskip 1em plus 0.5em minus 0.4em\relax New York, NY, USA: ACM,
  2018, pp. 50:1--50:6. [Online]. Available:
  \url{http://doi.acm.org/10.1145/3218603.3218615}
\BIBentrySTDinterwordspacing

\bibitem{Liang2018}
\BIBentryALTinterwordspacing
S.~Liang, S.~Yin, L.~Liu, W.~Luk, and S.~Wei, ``{FP-BNN: Binarized neural
  network on FPGA},'' \emph{Neurocomputing}, vol. 275, pp. 1072 -- 1086, 2018.
  [Online]. Available:
  \url{http://www.sciencedirect.com/science/article/pii/S0925231217315655}
\BIBentrySTDinterwordspacing

\bibitem{8884910}
C.~{Lammie}, W.~{Xiang}, and M.~R. {Azghadi}, ``{Accelerating Deterministic and
  Stochastic Binarized Neural Networks on FPGAs Using OpenCL},'' in \emph{2019
  IEEE 62nd International Midwest Symposium on Circuits and Systems (MWSCAS)},
  Aug 2019, pp. 626--629.

\bibitem{DBLP:journals/corr/abs-1812-11800}
\BIBentryALTinterwordspacing
S.~Darabi, M.~Belbahri, M.~Courbariaux, and V.~P. Nia, ``{BNN+: Improved Binary
  Network Training},'' \emph{CoRR}, vol. abs/1812.11800, 2018. [Online].
  Available: \url{http://arxiv.org/abs/1812.11800}
\BIBentrySTDinterwordspacing

\bibitem{AAAI1714619}
\BIBentryALTinterwordspacing
W.~Tang, G.~Hua, and L.~Wang, ``{How to Train a Compact Binary Neural Network
  with High Accuracy?}'' 2017. [Online]. Available:
  \url{https://aaai.org/ocs/index.php/AAAI/AAAI17/paper/view/14619}
\BIBentrySTDinterwordspacing

\bibitem{Lin2017}
\BIBentryALTinterwordspacing
X.~Lin, C.~Zhao, and W.~Pan, ``{Towards Accurate Binary Convolutional Neural
  Network},'' in \emph{Advances in Neural Information Processing Systems 30},
  I.~Guyon, U.~V. Luxburg, S.~Bengio, H.~Wallach, R.~Fergus, S.~Vishwanathan,
  and R.~Garnett, Eds.\hskip 1em plus 0.5em minus 0.4em\relax Curran
  Associates, Inc., 2017, pp. 345--353. [Online]. Available:
  \url{http://papers.nips.cc/paper/6638-towards-accurate-binary-convolutional-neural-network.pdf}
\BIBentrySTDinterwordspacing

\bibitem{DBLP:journals/corr/CaoL0Y17}
\BIBentryALTinterwordspacing
Z.~Cao, M.~Long, J.~Wang, and P.~S. Yu, ``{HashNet: Deep Learning to Hash by
  Continuation},'' \emph{CoRR}, vol. abs/1702.00758, 2017. [Online]. Available:
  \url{http://arxiv.org/abs/1702.00758}
\BIBentrySTDinterwordspacing

\bibitem{8461456}
C.~{Sakr}, J.~{Choi}, Z.~{Wang}, K.~{Gopalakrishnan}, and N.~{Shanbhag},
  ``{True Gradient-Based Training of Deep Binary Activated Neural Networks Via
  Continuous Binarization},'' in \emph{2018 IEEE International Conference on
  Acoustics, Speech and Signal Processing (ICASSP)}, April 2018, pp.
  2346--2350.

\bibitem{DBLP:journals/corr/abs-1902-00730}
\BIBentryALTinterwordspacing
F.~Lahoud, R.~Achanta, P.~M{\'{a}}rquez{-}Neila, and S.~S{\"{u}}sstrunk,
  ``{Self-Binarizing Networks},'' \emph{CoRR}, vol. abs/1902.00730, 2019.
  [Online]. Available: \url{http://arxiv.org/abs/1902.00730}
\BIBentrySTDinterwordspacing

\bibitem{DBLP:journals/corr/abs-1806-02988}
\BIBentryALTinterwordspacing
Z.~Li, D.~He, F.~Tian, W.~Chen, T.~Qin, L.~Wang, and T.~Liu, ``{Towards
  Binary-Valued Gates for Robust LSTM Training},'' \emph{CoRR}, vol.
  abs/1806.02988, 2018. [Online]. Available:
  \url{http://arxiv.org/abs/1806.02988}
\BIBentrySTDinterwordspacing

\bibitem{krizhevsky2009learning}
A.~Krizhevsky \emph{et~al.}, ``{Learning Multiple Layers of Features from Tiny
  Images},'' Citeseer, Tech. Rep., 2009.

\bibitem{Kingma2014AdamAM}
D.~P. Kingma and J.~Ba, ``{Adam: A Method for Stochastic Optimization},''
  \emph{CoRR}, vol. abs/1412.6980, 2014.

\bibitem{DBLP:journals/corr/abs-1805-07836}
\BIBentryALTinterwordspacing
Z.~Zhang and M.~R. Sabuncu, ``{Generalized Cross Entropy Loss for Training Deep
  Neural Networks with Noisy Labels},'' \emph{CoRR}, vol. abs/1805.07836, 2018.
  [Online]. Available: \url{http://arxiv.org/abs/1805.07836}
\BIBentrySTDinterwordspacing

\bibitem{pmlr-v38-lee15a}
\BIBentryALTinterwordspacing
C.-Y. Lee, S.~Xie, P.~Gallagher, Z.~Zhang, and Z.~Tu, ``{Deeply-Supervised
  Nets},'' in \emph{{Proceedings of the Eighteenth International Conference on
  Artificial Intelligence and Statistics}}, ser. Proceedings of Machine
  Learning Research, G.~Lebanon and S.~V.~N. Vishwanathan, Eds., vol.~38.\hskip
  1em plus 0.5em minus 0.4em\relax San Diego, California, USA: PMLR, 09--12 May
  2015, pp. 562--570. [Online]. Available:
  \url{http://proceedings.mlr.press/v38/lee15a.html}
\BIBentrySTDinterwordspacing

\bibitem{Simonyan2014VeryDC}
K.~Simonyan and A.~Zisserman, ``{Very Deep Convolutional Networks for
  Large-Scale Image Recognition},'' \emph{CoRR}, vol. abs/1409.1556, 2014.

\bibitem{pmlr-v15-glorot11a}
\BIBentryALTinterwordspacing
X.~Glorot, A.~Bordes, and Y.~Bengio, ``{Deep Sparse Rectifier Neural
  Networks},'' in \emph{Proceedings of the Fourteenth International Conference
  on Artificial Intelligence and Statistics}, ser. Proceedings of Machine
  Learning Research, G.~Gordon, D.~Dunson, and M.~Dudik, Eds., vol.~15.\hskip
  1em plus 0.5em minus 0.4em\relax Fort Lauderdale, FL, USA: PMLR, 11--13 Apr
  2011, pp. 315--323. [Online]. Available:
  \url{http://proceedings.mlr.press/v15/glorot11a.html}
\BIBentrySTDinterwordspacing

\bibitem{5457293}
J.~E. {Stone}, D.~{Gohara}, and G.~{Shi}, ``{OpenCL: A Parallel Programming
  Standard for Heterogeneous Computing Systems},'' \emph{Computing in Science
  Engineering}, vol.~12, no.~3, pp. 66--73, May 2010.

\bibitem{Sorensen:2016:HGC:2909437.2909440}
\BIBentryALTinterwordspacing
T.~Sorensen and A.~F. Donaldson, ``{The Hitchhiker's Guide to Cross-Platform
  OpenCL Application Development},'' in \emph{Proceedings of the 4th
  International Workshop on OpenCL}, ser. IWOCL '16.\hskip 1em plus 0.5em minus
  0.4em\relax New York, NY, USA: ACM, 2016, pp. 2:1--2:12. [Online]. Available:
  \url{http://doi.acm.org/10.1145/2909437.2909440}
\BIBentrySTDinterwordspacing

\bibitem{Krizhevsky:2017:ICD:3098997.3065386}
\BIBentryALTinterwordspacing
A.~Krizhevsky, I.~Sutskever, and G.~E. Hinton, ``{ImageNet Classification with
  Deep Convolutional Neural Networks},'' \emph{Commun. ACM}, vol.~60, no.~6,
  pp. 84--90, May 2017. [Online]. Available:
  \url{http://doi.acm.org/10.1145/3065386}
\BIBentrySTDinterwordspacing

\end{thebibliography}
\end{document}